\documentclass{ecai} 

\usepackage[most]{tcolorbox}
\usepackage{subfigure}
\usepackage{latexsym}
\usepackage{amssymb}
\usepackage{amsmath}
\usepackage{amsthm}
\usepackage{booktabs}
\usepackage{enumitem}
\usepackage{graphicx}
\usepackage{color}

\newcommand{\BibTeX}{B\kern-.05em{\sc i\kern-.025em b}\kern-.08em\TeX}

\begin{document}


\begin{frontmatter}


\paperid{6171} 


\title{Disentangling Exploration of Large\\ Language Models by Optimal Exploitation}


\author[A]{\fnms{Tim}~\snm{Grams}\thanks{Corresponding Author. Email: tim.nico.grams@tu-clausthal.de.}}
\author[B]{\fnms{Patrick}~\snm{Betz}}
\author[B]{\fnms{Sascha}~\snm{Marton}}
\author[C]{\fnms{Stefan}~\snm{Lüdtke}}
\author[A]{\fnms{Christian}~\snm{Bartelt}}

\address[A]{Technical University of Clausthal}
\address[B]{University of Mannheim}
\address[C]{University of Rostock}


\begin{abstract}
Exploration is a crucial skill for in-context reinforcement learning in unknown environments. However, it remains unclear if large language models can effectively explore a partially hidden state space. This work isolates exploration as the sole objective, tasking an agent with gathering information that enhances future returns. Within this framework, we argue that measuring agent returns is not sufficient for a fair evaluation. Hence, we decompose missing rewards into their exploration and exploitation components based on the optimal achievable return. Experiments with various models reveal that most struggle to explore the state space, and weak exploration is insufficient. Nevertheless, we found a positive correlation between exploration performance and reasoning capabilities. Our decomposition can provide insights into differences in behaviors driven by prompt engineering, offering a valuable tool for refining performance in exploratory tasks.
\end{abstract}

\end{frontmatter}


\section{Introduction}
\textbf{Large language models in decision making.} Recently, frontier large language models (LLM) demonstrated promising results in various decision-making tasks such as web browsing \cite{yao2022react, shinn2024reflexion, ma2023laser}, game-playing \cite{paglieri2024balrog}, and simulated households \cite{yao2022react, shinn2024reflexion}. Hereby, LLMs act as agents that observe states and take actions for in-context reinforcement learning (RL) in unknown environments. Through their vast internal knowledge base and autoregressive reasoning capabilities, the models are supposed to quickly adapt to new tasks. However, prior work has shown that LLMs struggle with complex problems due to several shortcomings: For example, the ability to learn from mistakes \cite{huang2023selfcorrect} and to plan over long horizons is often limited \cite{kambhampati2024planning}. 

\textbf{Exploration is essential for self-improvement.} In many tasks, an agent's ability to self-improve relies on a sufficient exploration of the environment. Exploration is the skill of seeking novel information to reduce uncertainty and improve decision-making over time. Traditional approaches often rely on stochastic noise \cite{mnih2013playing, lillicrap2015continuous} or reward-shaping \cite{bellemare2016unifying}. However, such methods induce redundancy and differ from human decision-making, which is mainly driven by reasoning about current information, past steps, and the potential of yet unknown actions. Therefore, recent works proposed more intelligent approaches by incorporating the reasoning capabilities of LLMs \cite{huang2024wese, lu2024intelligent, nie2024evolve}. LLMs could transform exploration into a targeted and efficient process by identifying patterns, leveraging prior knowledge, and intelligently navigating the state space.

\textbf{Limited focus on exploration as an independent ability.}
While exploration is acknowledged as a crucial component for learning and decision making, prior work has investigated it in conjunction with exploitation \cite{krishnamurthy2024can, nie2024evolve, paglieri2024balrog, ke2024interactive, chen2024efficient}. 
In this context, exploration is often measured indirectly through cumulative rewards or success rates. 
However, this conflates the ability with overall performance. It is difficult to isolate an agent’s success in exploration, as true progress is independent of the agent’s return. 
While seeking information, agents are supposed to sacrifice short-term rewards in favor of long-term understanding. 
Also, strong exploration does not always translate into high performance if the agent lacks the ability to exploit the knowledge it has gathered.  An agent may thoroughly explore an environment but fail to leverage its findings. 
Therefore, it is essential to develop methods that accurately measure exploration of LLMs, as it helps to foster our understanding of their learning and adaptability in sequential decision making.

\begin{figure*}[t!]
    \centerline{\includegraphics[width=0.75\textwidth]{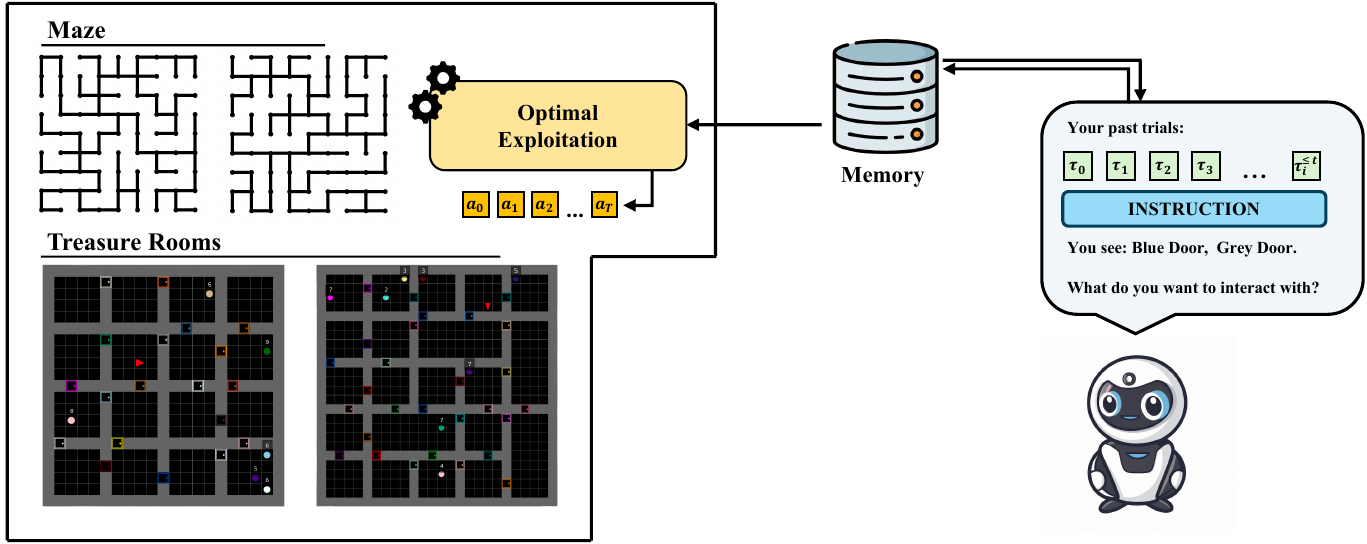}}
    \caption{\textbf{An overview of our evaluation procedure based on an optimal exploitation}. The agent navigates a maze or grid of rooms by interacting with doors and balls. Overall, the goal is to gather information about highly rewarding balls. After each episode, we calculate the optimal sequence of actions to measure exploration progress based on the agent's current memory.}
    \label{fig:method_overview}
\end{figure*}
\textbf{Contribution.} In this paper, we claim that exploration will be an essential skill for future agents. Recent studies \cite{krishnamurthy2024can, nie2024evolve} investigated LLMs balancing exploration and exploitation, where they found that they often struggle to achieve high rewards. 
Conversely, we disentangle exploration for an independent evaluation. 
In this controlled setting, we argue that conventional metrics, such as agent returns, LLM exploitation, or the convergence towards the best sequence of actions \cite{huang2024wese, krishnamurthy2024can} are insufficient.
To fill this gap, we make the following contributions: First, we propose to measure the \emph{optimal exploitation return} (see Figure \ref{fig:method_overview}) and decompose missing rewards into their exploration and exploitation components. This decomposition measures exploration progress accurately and enables insights into reward sacrifice. Afterward, to the best of our knowledge, we conduct the first comprehensive evaluation of various popular closed and open-source LLMs with the exploration being disentangled. 

\textbf{Results.} Our findings indicate that most LLMs face challenges with disentangled state space exploration. Our results align with prior work \cite{krishnamurthy2024can, paglieri2024balrog}, which highlights the trade-off between exploration and exploitation. Further, we observe that state space coverage diminishes significantly when agents are required to plan over extended horizons, indicating a limitation in their long-term exploratory capabilities. Contrary to earlier claims \cite{huang2024wese}, our analysis demonstrates that weak LLMs are insufficient. Notably, we identified a significant trend wherein stronger exploration capabilities correlate with language understanding and reasoning abilities, suggesting that more advanced LLMs may possess enhanced exploration potential. 
\\
\\Besides that, we found that our method can be used to uncover model-specific differences in reward sacrifice. In this context, we demonstrate how to assess the impact of agent instructions on exploratory behavior, providing insights for prompt engineering.

\section{Related Work}
Recently, exploring the capabilities of LLMs has gained significant attention and has been investigated from different perspectives. Prior work includes, but is not limited to, planning \cite{kambhampati2024planning, valmeekam2023planning}, world modeling \cite{hao2023worldmodel}, logical- \cite{xu2023logical}, commonsense- \cite{li2021systematic}, and mathematical reasoning \cite{imani2023mathprompter, yuan2023scaling}. Our research focuses on an intelligent exploration of the state space in sequential decision making. We summarize existing methods into three streams of work: 

\textbf{Large language models in decision making.}
A broad line of work has applied LLMs in decision-making. Relatively few addressed the challenge of exploration. The majority of existing research focuses on multi-armed bandit problems \cite{park2024regret, krishnamurthy2024can, nie2024evolve, chen2024efficient}. For example, in \citet{krishnamurthy2024can} the action space consists of buttons with different stochastic rewards. At each timestep, the LLM is presented with a history of past interactions and must maximize the return over multiple trials. 
Other research examined exploration in hypothesis testing \cite{piriyakulkij2024doing, ke2024interactive}. Only a limited number of works \cite{ma2023laser, paul2023sequential, huang2024wese, lu2024intelligent} proposed methods for sequential state spaces. \citet{lu2024intelligent} suggest using an LLM to measure the interestingness of states to return to. \citet{huang2024wese} employ small-scale models for exploration and larger ones for exploitation. Generally, evaluations focus on agent returns. While LLMs are used for exploration, an isolated evaluation is missing. Therefore, our study is specifically on LLMs that perform information gathering about their environment. 

\textbf{Intrinsic motivation and reinforcement learning.}
Our evaluation is closely related to exploration in RL.
Early approaches predominantly relied on random methods, such as epsilon-greedy strategies \cite{mnih2013playing, lillicrap2015continuous}. More recent research has shifted toward developing intelligent exploration techniques \cite{ecoffet2019go, norman2023first, mazoure2023accelerating} and extending algorithms with intrinsic motivation \cite{schmidhuber2010formal, klyubin2005empowerment, pathak2017curiosity, bellemare2016unifying, achiam2017surprise, burda2018exploration}. Methods that decouple exploration and exploitation are particularly relevant to our work \cite{avner2012decouplingbandit, norman2023first, schafer2021decoupling, whitney2021decoupled}. For example, First-Explore \cite{norman2023first} uses separate actors for exploration and exploitation, with the exploration policy informing subsequent exploitation decisions. Generally, RL focuses on stabilizing the decoupled policies. Evaluations use non-optimal exploitation returns as optimal policies can be hard to obtain in complex environments. On the contrary, we interpret an optimal exploitation as a diagnostic tool, analogous to oracle policies and optimal value functions in the literature. We make use of environments with pre-specified optimal exploitation algorithms, as agent returns and LLM exploitation do not provide reliable estimates. 

\textbf{Evaluation environments.}
A variety of environments has been developed to assess different aspects of LLMs. 
Notable examples that require exploration include TextWorld \cite{cote2019textworld}, Crafter \cite{hafner2021benchmarking}, NetHack \cite{kuttler2020nethack}, MiniHack \cite{samvelyan2021minihack}, MiniGrid \cite{chevalier2024minigrid}, and BabyAI \cite{chevalier2018babyai}. There is also growing interest in environments for computer and web agents \cite{zhou2023webarena, yao2022webshop}. Comprehensive benchmarks, including SmartPlay \cite{wu2023smartplay}, AgentBench \cite{liu2023agentbench}, LMAct \cite{ruoss2024lmact}, and BALROG \cite{paglieri2024balrog}, aggregate multiple environments to evaluate agents across a wide range of capabilities. In some of them, \citet{paglieri2024balrog} observed aimless movements and redundant behaviors. 
The difference in our work lies in the isolation of exploration. Our experiments are based on mazes and an adapted version of Dark Treasure Room \cite{norman2023first}. Our environments offer limited prior knowledge, provide non-binary, relatively dense rewards, and maintain a high-level, symbolic action space. 

\section{Evaluating with Optimal Exploitation}
In the following section, we present our evaluation framework. We introduce the formal background, an optimal exploitation oracle as a diagnostic tool for a fair comparison, and present our regret decomposition. Additionally, we introduce Maze and Treasure Rooms environments, which provide dense rewards and an, at this point, unsolved challenge for instruction-tuned, frontier LLM exploration in sequential decision-making. 

\subsection{Theoretical Framework}
\label{sec:theoretical_framework}
\textbf{Markov Decision Process.}
We study a deterministic Markov decision process (MDP) $(\mathcal{S}, \mathcal{A}, \mathcal{P}, r, \gamma)$ where $\mathcal{S}$ is a finite state space, $\mathcal{A}$ is the finite action space, $\mathcal{P}: \mathcal{S} \times \mathcal{A} \times \mathcal{S} \rightarrow [0,1]$ defines the transition dynamics of the environment, $r: \mathcal{S} \times \mathcal{A} \rightarrow \mathbb{R}$ is the reward function and $\gamma$ the discount factor. At each time step $t$, an agent samples an action from policy $\pi: \mathcal{S} \rightarrow \mathcal{A}$ based on the current observation $s_t \in \mathcal{S}$ and executes it in the environment. The environment transitions and the agent receives a reward $r_t$. In an episodic MDP, the procedure repeats until a terminal state $s_T$ is reached or the maximum number of steps is exceeded. The sequence of state, action, and reward triples until state $s_T$ is called a trajectory and describes the interaction within an episode $\tau_i = \left( s_0, a_0, r_0, s_1, a_1, r_1, s_2, \dots \right)$.\linebreak[4] In our in-context RL setting, the agent has access to the history of the current and prior episode interactions $h_{i,t} = (\tau_1, \dots, \tau_{i-1}, \tau_i^{\leq t})$, where $\tau_i^{\leq t} = (s_0,a_0,r_0,\dots,s_t)$ denotes the partial trajectory of ongoing episode i. The goal is to maximize the cumulative return $R_i = \sum_{t=0}^{T} r_t$. 
An alternative way of measuring progress and success is the agent's cumulative regret, which quantifies the loss in return due to not always picking the best available action. Let $a_t^* = \arg\max_{a} \mathbb{E}[r_t \mid s_t, a]$ be the optimal action at step $t$ given the current state $s_t$ and let $r_t^*$ denote the reward obtained from executing $a_t^*$. Similarly, let $a_t$ be the action chosen by the agent, yielding reward $r_t$. The cumulative regret over T steps is then defined as
\begin{equation}
    \text{Regret}_i(T) = \sum_{t=0}^{T} \left[ r^*_t - r_t \right]
\end{equation}

\textbf{Exploitation versus exploration.}
The cumulative regret can be influenced by an agent's ability of exploitation and exploration. For \emph{exploitation}, the agent acts greedily. 
Given history $h_{i,t}$, the agent samples actions $a_t \sim \pi_{\text{exploit}}(\cdot \mid h_{i,t})$ from a policy concentrated around the greedy choice $\pi_{\text{exploit}}(h_{i,t}) \approx \arg\max_{a \in \mathcal{A}} \mathbb{E}\!\left[r(s_t,a) \mid h_{i,t}\right]$, thus aiming to maximize the immediate return. Similarly, the exploitation policy utilizes history $h_{i,t}$ to estimate the action sequence that yields the lowest regret. 
 On the contrary, \emph{exploration} emphasizes interacting with the environment to sacrifice short-term rewards for information, which can reduce uncertainty and improve future decision-making. A common objective is to take non-greedy actions to discover trajectories that minimize the regret at subsequent exploitation episodes. Thus, exploration should sample action $a_t \sim \pi_{\text{explore}}(\cdot \mid h_{i,t})$ from a policy $\pi_{\text{explore}} \approx \arg\max_{\pi_{\text{explore}}} \mathbb{E} \left[ R^{\pi_{\text{exploit}}}_{i+1} \mid h_{i+1} = h_i \cup \tau_i \right]$ to add $r_i$ to prior trajectories $h_i$ such that return $R_{i+1}^{\pi_{\text{exploit}}}$ of $\pi_{\text{exploit}}$ is maximal. 

\subsection{Disentangling Exploration}
\begin{figure}[t]
    \centering
    \includegraphics[width=0.75\linewidth]{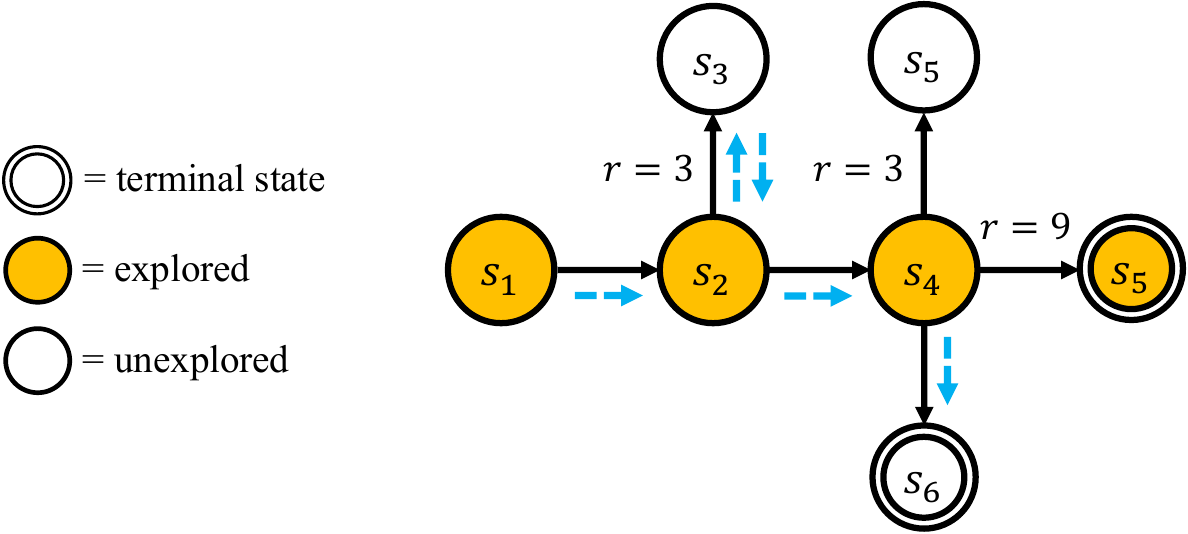}
    \caption{\textbf{Exploration is independent of the episode return.} Discovering a trajectory (depicted as blue arrows) over $s_3$ to $s_6$ does not immediately improve return ($R=3$). However, an optimal exploitation displays progress appropriately $(R^{\sigma_{\text{exploit}}}=12)$.}
    \label{fig:main-paper-mdpp}
\end{figure}

\textbf{The necessity of optimal exploitation.} When an agent does both exploitation and exploration, progress can be evaluated by disabling exploration and measuring the return of the agent at the end of an episode. Nevertheless, when an agent is concerned purely with exploration, the return alone does not accurately reflect success, as good exploration does not necessarily yield higher returns. Consider the MDP in Figure \ref{fig:main-paper-mdpp} as an example and let $R_i$ and $R_{i+1}$ be an episode of exploration followed by exploitation. Then, observing a trajectory over $s_3$ to $s_5$ does not yield a higher return during exploration ($R_i=3$), but may improve exploitation ($R_{i+1}=12$) in episode $i+1$. However, we found measuring exploitation returns with an LLM to be similarly hard to exploration without giving reliable estimates (see Appendix A.7 \cite{grams2025disentangling}).

\textbf{Exploitation oracles for an accurate exploration metric.}
Due to the above-observed discrepancy between $R_i$ and $R_{i+1}$, we introduce an optimal exploitation oracle $\sigma_{\text{exploit}}$ for an accurate isolation of an agent's exploration performance. In this context, the optimal exploitation oracle is a predefined policy that, given the agent’s history, optimally selects the locally reward-maximizing action $a_t \sim \sigma_{\text{exploit}}(s_t, h_i)$ at every time step, and can be interpreted as a diagnostic tool analogous to oracle policies and optimal value functions in the literature. 
The optimal exploitation allows an accurate disentanglement of exploration. 
\textbf{Why?} If the return under optimal exploitation is not as high as the maximal achievable return, the only reason can be a lack of exploration. 
Hence, by calculating the optimal exploitation return, we can assess an agent’s exploration capabilities accurately without conflating them with exploitation. 

\textbf{Decomposing the cumulative regret.}
Let $R^{\sigma_{\text{exploit}}}_i$ be the return of oracle $\sigma_{\text{exploit}}$ and $R^*_i$ the maximal achievable return in episode $i$. Based on $R^{\sigma_{\text{exploit}}}_i$, we can then decompose the cumulative regret into its exploitation and exploration components. First, we start by defining the exploration gap
\begin{equation}
    \Delta^{\text{explore}}_i = R^*_i - R^{\sigma_{\text{exploit}}}_i
\end{equation}
which quantifies the loss in return due to missing information. As mentioned previously, the only cause of uncertainty with an optimal exploitation oracle can be a lack of exploration. Therefore, $\Delta^{\text{explore}}_i$ measures true exploration progress. 
Finally, we define the exploitation gap $\Delta^{\text{exploit}}_i$ as the difference between the return of the optimal exploitation oracle and the agent's return
\begin{equation}
    \Delta^{\text{exploit}}_i = R^{\sigma_{\text{exploit}}}_i - R_i
\end{equation}
which measures the cost for exploring the environment to find future information instead of optimally exploiting the information of the current history. 
Note that with deterministic environment transitions and reward functions, it follows that $R^*_i \geq R^{\sigma_{\text{exploit}}}_i \geq R_i$. Also, it is easy to see that it holds by construction:
\begin{equation}
    \text{Regret}_i(T) = \Delta^{\text{explore}}_i + \Delta^{\text{exploit}}_i.
\end{equation}

\textbf{A synthetic example.} To illustrate the value of our framework, we refer again to the example MDP in Figure \ref{fig:main-paper-mdpp}. At episode $i$, the $\text{Regret}_i(T)=6$ shows missing rewards purely due to insufficient exploration $\Delta^{\text{explore}}_i=6\ (100\%)$. However, during subsequent exploration, the agent discovers the additional transition from $s_3$ to $s_4$, which provides information about the reward distribution. Nevertheless, the agent sacrificed short-term return for this knowledge. Our framework accurately captures that discrepancy by decomposing the $\text{Regret}_{i+1}(T)=12$ into its exploration $\Delta_{i+1}^{\text{explore}}=3\ (25\%)$ and exploitation components $\Delta_{i+1}^{\text{exploit}}=9\ (75\%)$. Our metric shows that the agent made exploration progress as the exploration gap lowered, although cumulative regret increased. 

\textbf{On the complexity of defining an optimal exploitation.}
While the possibility of deriving optimal exploitation is a strong assumption, it is necessary to disentangle exploration from exploitation accurately. We argue that this assumption is acceptable when focusing on a correct evaluation of exploration capabilities. 
Theoretically, an optimal exploitation policy exists for most environments, and our approach could be applied universally. In practice, finding the optimal exploitation is intractable in many environments.
Defining the optimal exploitation oracle for evaluation purposes upfront with full knowledge about the environment eases the assumption in many cases. Approximators are an option, but should be consistent; otherwise, they might introduce ambiguity. 

\subsection{Environments and Optimal Exploitation}
To mitigate the impact of low-level spatial navigation, we design environments with a higher-level, symbolic action space, which is similar to TextWorld \cite{cote2019textworld}. Each environment has a fixed layout and starting position between episodes. Episodes terminate when three objects are collected or a maximum number of transitions is exceeded. The latter is set to the distance between the starting position and the furthest room, ensuring that the budget scales with complexity. 

\textbf{Treasure Rooms.}
The first task follows a grid of rooms and is an adapted version of Dark Treasure Room \cite{norman2023first}. Each map contains balls which are placed randomly on the grid and associated with a constant reward $\in[1, 10] \cap \mathbb{Z}$. The agent’s goal is to maximize the return by collecting at most three balls within the step budget. As reward locations and their values are drawn upfront, the optimal exploration strategy, across multiple layouts, requires full state space coverage.
An exemplary agent-environment interaction is depicted in Figure \ref{fig:method_overview} and all environments are illustrated in Appendix A.1 \cite{grams2025disentangling}. 

\textbf{Maze.}
Additionally, we compare exploration in random mazes to scale beyond simple grid layouts. Mazes are common in RL literature on in-context learning abilities \cite{laskin2022context, grigsby2023amago, schmied2024retrieval}. Again, we take inspiration from \citet{norman2023first} and place multiple rewards in the environment. In contrast to Treasure Rooms, rewards are more sparsely distributed. Further, mazes demand increased long-horizon planning as early decisions have more impact on possible future paths. The maze layouts can be found in Appendix A.2 \cite{grams2025disentangling}.  

\textbf{Optimal exploitation oracles.}
The optimal exploitation oracle is fastest by solving an instance of the orienteering optimization problem that searches for the path between balls with the highest reward. See more details in Appendix A.6 \cite{grams2025disentangling}. Generally, our framework is independent of the procedure of optimal exploitation. In future work, we envision imitation learning techniques on the agent's history as a scalable method to find actions for an optimal exploitation.  

\section{Experimental Evaluation} \label{sec:experiment-eval}
Our experiments aim to demonstrate the utility of our evaluation method by addressing the following research questions: 1.) Can LLMs explore and reduce the exploration gap? and 2.) How do agent instructions impact exploration and exploitation gaps?

\subsection{Setup} \label{sec:setup}
\textbf{Agent instruction.} Our experimental setup directs agents to explore and gather information about the balls. We reference this prompt as \emph{task-oriented} exploration. The exact template is detailed in Appendix A.4 \cite{grams2025disentangling}. Each agent is equipped with a simple memory that stores the history of past interactions. Unlike many prior studies, our approach does not provide agents with explicit hints about the domain. 

\textbf{Statistical robustness.} We test each open-source with ten runs and each closed model with five runs on every environment. Further, we employ paired t-tests for metric comparisons between models. In the following Figures, we report statistical significance using the standard error.
Our return decomposition is measured at the end of exploration and as the mean over all episodes. The values are normalized to account for environments with different maximum returns. Full results are listed in Appendices B and C \cite{grams2025disentangling}. 

\textbf{Evaluation procedure.} We evaluate our first research question across a total of four different sizes of Treasure Rooms and maze environments. The configurations are designed to test the scalability of the planning horizon by gradually increasing the environment size.\linebreak[4] For our second research question, we compare the task-oriented agent instruction with alternative prompts in a 7$\times$7 grid. Specifically, we design prompts that are supposed to test lower and upper bounds for the exploration capabilities by directing the LLM to exploit or explore intrinsically without referencing the task. See Appendix D \cite{grams2025disentangling} for more information and the alternative instructions.

\subsection{Other Statistics}
We also describe exploration using the following statistics: 

\textbf{Agent return}
represents the cumulative rewards collected by the agent. Although it is not a direct indicator of effective exploration, this metric reflects the ability to engage in ways that yield task success. A higher agent return implies that the agent is possibly close to exploitation. Contrary to our decomposition, it does not necessarily imply that the agent is effectively exploring.

\textbf{State space coverage}
reflects the ability to discover novel states. In our experiments, effective exploration requires covering the entire state space, as balls with high rewards are randomly spread. However, complete exploration may not always be necessary or can be infeasible in large, infinite, or continuous spaces. Our decomposition also covers cases in which full exploration is not essential. 

\textbf{Memory redundancy}
measures the fraction of redundant state-action pairs in the memory. In our environments, revisiting states or actions does not provide new information. Reducing redundancy conserves resources and ensures that the history can be utilized efficiently. Otherwise, information has to be retrieved from longer histories to find novel or informative states to explore.

\textbf{Sample efficiency}
evaluates the number of environment interactions required to converge to the 90\% equilibrium of the maximum achieved exploitation return. Note that sample efficiency is decoupled from the quality of the final policy. For example, constantly stuck with the same low-reward balls is equally valued as fast convergence to the maximum exploitation return. 

\subsection{Models} 
We evaluate Mistral (7B) \cite{jiang2023mistral}, Gemma2 (9B and 27B) \cite{team2024gemma}, Llama 3.1 (7B) \cite{dubey2024llama}, Llama3.3 (70B), Mistral Small (22B) \cite{mistralAbundance}, and Mistral Large 2 (123B) \cite{mistralLargeEnough}. We further include closed-models GPT-4o \cite{achiam2023gpt}, o1-mini \cite{OpenAIo1}, Gemini 1.5 Pro \cite{team2024gemini}, and Claude 3.5 Sonnet \cite{anthropicIntroducingClaude}. 
We compare the LLMs to a policy that takes random actions. 

\subsection{Can Large Language Models Explore?}
\label{sec:completeness}
\textbf{Agents struggle to minimize the exploration gap.}
The results of our experiments indicate that most LLMs have difficulty exploring the state space, also when exploration for information gathering has been disentangled. Average results are presented in Table \ref{average-results-gaps} and \ref{average-results-others}. Almost all models demonstrate significantly larger or indifferent exploration gaps in their final episodes compared to the random baseline. Only Mistral Large, o1-mini, and Gemini 1.5 Pro achieve smaller exploration gaps. However, all failed to achieve complete state space coverage, exhibiting non-zero exploration gaps and exploring less than 90\% of the state space. Additionally, performance was significantly worse in the maze environment, where the increased structural complexity and sparse reward distribution led to lower coverage and larger exploration gaps. At closer inspection, LLMs showed redundant visitation of previous states, disproportionately focusing on initial high-reward regions. 

\begin{figure*}[t]
    \centering
    \includegraphics[height=0.24\textwidth]{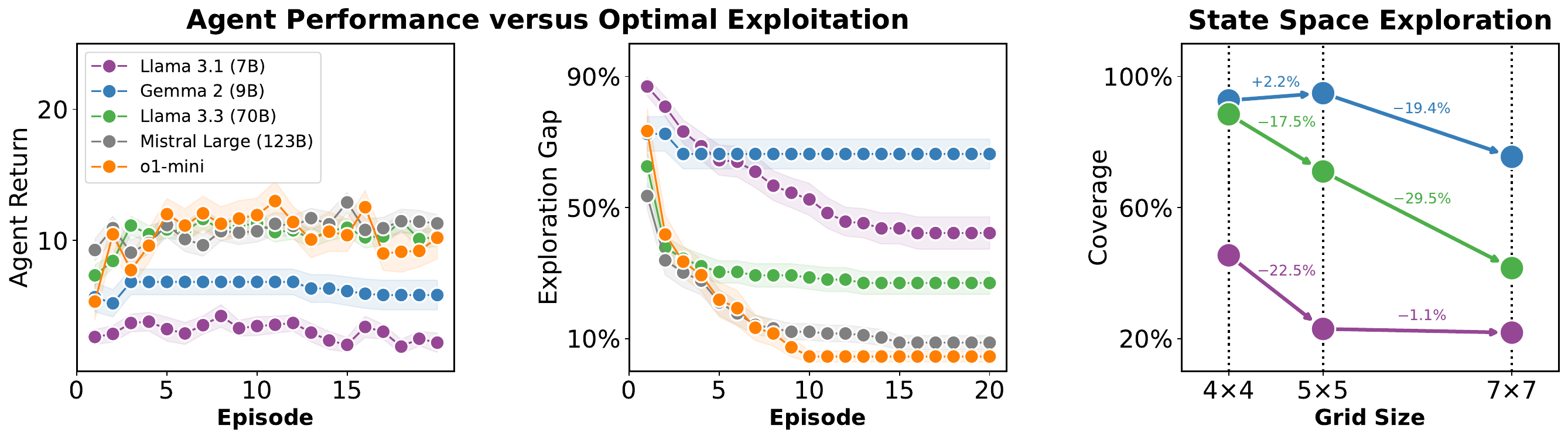}
    \caption{\textbf{The agent return, exploration gap, and grid coverage in $5 \times 5$ Treasure Rooms.} Our exploration gap (middle, lower is better) measures progress in exploration accurately, in contrast to the agent return. Note, for example, that Llama3.1 (7B) performs better exploration while achieving lower agent return than Gemma 2 (9B). Also, Llama 3.3 (70B) and o1-mini do not show significant differences in the agent return, although they are significantly different (p $\leq 0.01$) in their exploration gap. Our experiments further demonstrate that, despite being given sufficient trials, coverage shrinks with increased long-horizon planning.}
    \label{fig:main-paper-learning-curves}
\end{figure*}
\begin{table*}[t]
\caption{\textbf{Regret decomposition averaged over all variations of Treasure Rooms.} The exploitation and exploration gaps are normalized to account for varying maximum rewards in different environments. The numbers in parentheses show the percentage that exploitation and exploration gaps make up of the cumulative regret, respectively.}
\label{average-results-gaps}
\centering
\resizebox{0.8\textwidth}{!}{%
\begin{tabular}{ll||ccc||ccl}
\hline
 &  & \multicolumn{3}{c||}{\textbf{Last Episode}} & \multicolumn{3}{c}{\textbf{Mean}} \\ \hline
\multicolumn{1}{c}{\textbf{}} & \multicolumn{1}{c||}{\textbf{}} & \textbf{Regret} & $\Delta^{\text{exploit}}$ & $\Delta^{\text{explore}}$ & \textbf{Regret} & $\Delta^{\text{exploit}}$ & $\Delta^{\text{explore}}$ \\ \hline
\textbf{\textless 30B} & Mistral (7B) & 0.80 & 0.25 (31\%) & 0.55 (69\%) & 0.78 & 0.20 (26\%) & 0.57 (74\%) \\
 & Llama 3.1 (7B) & 0.84 & 0.35 (41\%) & 0.49 (59\%) & 0.82 & 0.24 (28\%) & 0.57 (72\%) \\
 & Gemma 2 (9B) & 0.75 & \textbf{0.15} (17\%) & 0.60 (83\%) & 0.69 & \textbf{0.08} (10\%) & 0.60 (90\%) \\
 & Mistral Small (22B) & 0.68 & 0.28 (41\%) & 0.40 (59\%) & 0.69 & 0.20 (29\%) & 0.49 (71\%) \\
 & Gemma 2 (27B) & \textbf{0.57} & 0.18 (32\%) & \textbf{0.38} (68\%) & \textbf{0.52} & 0.12 (22\%) & \textbf{0.40} (78\%) \\ \hline
\textbf{\textgreater 30B} & Llama 3.3 (70B) & 0.50 & \textbf{0.23} (40\%) & 0.27 (60\%) & 0.50 & \textbf{0.18} (32\%) & 0.32 (68\%) \\
 & Mistral Large (123B) & \textbf{0.33} & 0.26 (75\%) & \textbf{0.07} (25\%) & \textbf{0.38} & 0.25 (62\%) & \textbf{0.13} (38\%) \\ \hline
\textbf{Closed} & GPT-4o & 0.44 & 0.14 (31\%) & 0.30 (69\%) & 0.49 & 0.14 (27\%) & 0.35 (73\%) \\
 & GPT-o1-mini & 0.43 & 0.38 (79\%) & \textbf{0.06} (21\%) & 0.47 & 0.32 (66\%) & \textbf{0.16} (34\%) \\
 & Gemini 1.5 Pro & 0.52 & 0.30 (58\%) & 0.22 (42\%) & 0.53 & 0.23 (41\%) & 0.30 (59\%) \\
 & Claude 3.5 Sonnet & \textbf{0.38} & \textbf{0.09} (29\%) & 0.29 (71\%) & \textbf{41\%} & \textbf{0.09} (26\%) & 0.32 (74\%) \\ \hline
\textbf{Baseline} & Random Walk & 0.85 & 0.53 (62\%) & 0.32 (38\%) & 0.86 & 0.37 (42\%) & 0.49 (58\%) \\ \hline
\end{tabular}%
}
\end{table*}

\textbf{Limitations in targeted, long-horizon exploration.}
The fraction of the explored state space in Figure \ref{fig:main-paper-learning-curves} (right) reveals a significant decrease ($p\leq0.01$) in explored states as the environment in Treasure Rooms gets larger. Models with more parameters, including GPT-4o, Mistral Large, and Claude 3.5 Sonnet, maintained consistent coverage only when the environment size increased slightly ($p\geq0.05$).
The reasoning model o1-mini experienced a similar performance drop, despite achieving superior results in smaller environments. However, we leave a more thorough analysis of reasoning models for future work. All models exhibited the largest exploration gaps in the maze environments, underscoring the limitations of careful long-horizon planning in more complex settings.

\begin{figure*}[t!]
    \centering
    \includegraphics[width=0.6\textwidth]{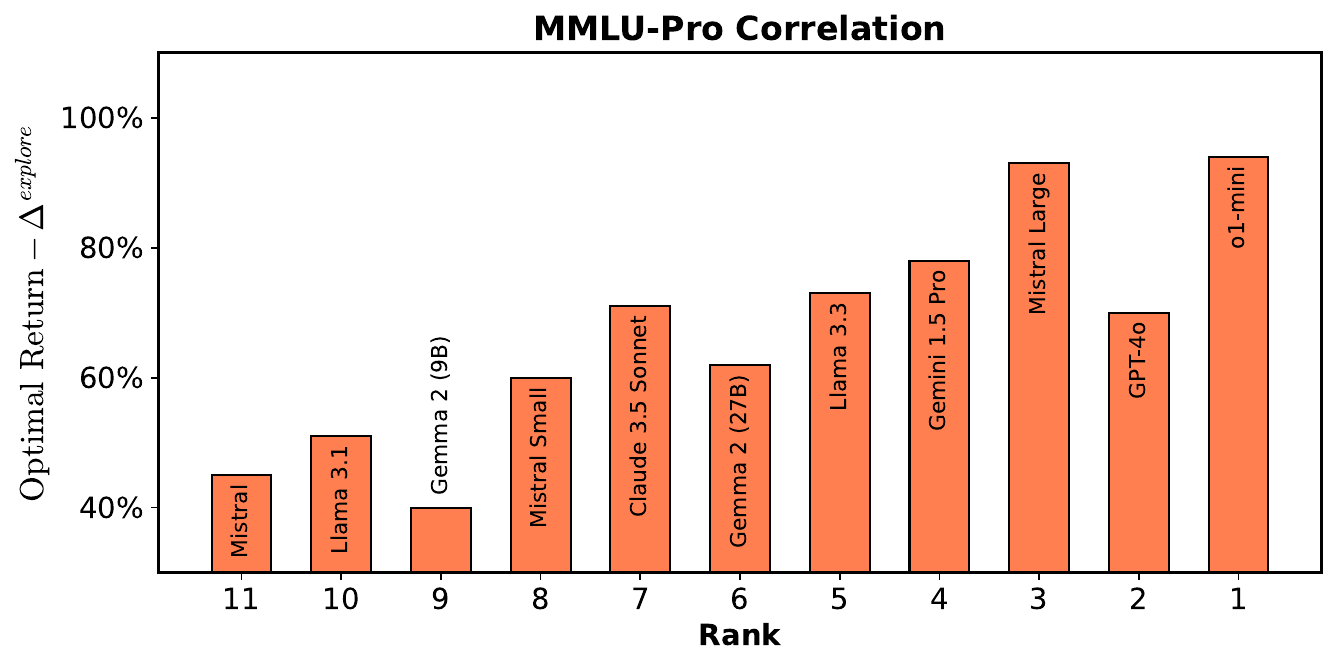}
    \caption{\textbf{Exploration gaps strongly correlate with knowledge and reasoning capabilities.} We found that models achieving high scores in MMLU-Pro are more competent in exploring their environment. We test this hypothesis by regressing the average exploration gaps in Treasure Rooms and the mazes with the results of the MMLU-Pro benchmark. A linear regression confirmed a significant correlation ($p \leq 0.01$). These outcomes indicate that less capable models are not more uncertain or associated with sampling different paths. Instead, produce lower-quality solutions and stick to the same trajectories. }
    \label{fig:main-paper-correlation}
\end{figure*}

\textbf{Weak exploration is not enough.}
In prior work, small language models were sufficient for exploration \cite{huang2024wese}. Motivated by these claims, we correlated the exploration gaps with each LLM's result in the Massive Multilingual Language Understanding (MMLU-Pro) benchmark \citep{wang2024mmlu}. Contrary to earlier claims, we found that specifically weak LLMs struggle to explore their environment effectively. We hypothesize the difference is due to the disentangled evaluation and increased exploration complexity compared to prior experiments.
Figure \ref{fig:main-paper-correlation} reveals a clear trend of improved exploration capabilities with MMLU-Pro score, as evidenced by the statistically significant slope of a linear regression ($p\leq0.01$).
Interestingly, closed-source models do not necessarily outperform open-source counterparts. For instance, GPT-4o and Llama 3.3 exhibit comparable exploration gaps at the end of training. 

\textbf{The regret decomposition measures exploration accurately.}
Our results support the proposed regret decomposition to obtain exploitation and exploration components. To support our claim, we refer the reader again to the agent returns and the corresponding exploration gaps, which are illustrated in Figure \ref{fig:main-paper-learning-curves}. We find that the agent's return often does not correlate with true exploration progress. For instance, Llama 3.3 (70B) and o1-mini exhibit no significant differences in their agent returns, despite showing statistically significant differences in their exploration gaps ($p\leq0.01$). 
To test whether we can also use an LLM as a replacement for our optimal exploitation, we conducted an ablation study in which we substituted our optimal exploration with an Llama 3.3 (70B) and compared their results. The outcomes for this follow-up experiment are listed in Appendix A.7 \cite{grams2025disentangling}. The additional ablation demonstrates that the exploitation return cannot be reliably measured in this setting.  These findings underscore our hypothesis that exploitation and exploration are equally challenging, and they should be assessed independently to ensure an accurate evaluation. 

\subsection{What is the Impact of Agent Instructions?}
\begin{table*}[t]
\caption{\textbf{The other statistics averaged over all variations of Treasure Rooms.} Some LLMs achieve lower agent returns but are higher in their optimal exploitation return (see Gemini 1.5 Pro and Claude 3.5 Sonnet). We find that all agents converge quickly to their individual maximum optimal exploitation return, although more rewarding balls are not explored yet.}
\label{average-results-others}
\centering
\resizebox{0.88\textwidth}{!}{%
\begin{tabular}{ll||ccccc}
\hline
\multicolumn{1}{c}{\textbf{}} & \multicolumn{1}{c||}{} & \textbf{Exploitation Return} & \textbf{Agent Return} & \textbf{Coverage} & \textbf{Redundancy} & \textbf{Sample Efficiency} \\ \hline
\textbf{\textless 30B} & Mistral (7B) & 9.47 & 4.13 & 38.88\% & 92.17\% & 13.82\% \\
 & Llama 3.1 (7B) & 10.56 & 3.27 & 37.03\% & 92.25\% & 31.25\% \\
 & Gemma 2 (9B) & 8.53 & 5.18 & 30.10\% & 95.25\% & \textbf{7.25\%} \\
 & Mistral Small (22B) & 12.54 & 6.67 & 46.19\% & \textbf{91.43\%} & 35.44\% \\
 & Gemma 2 (27B) & \textbf{12.82} & \textbf{8.99} & \textbf{49.99\%} & 92.43\% & 14.13\% \\ \hline
\textbf{\textgreater 30B} & Llama 3.3 (70B) & 14.94 & 10.32 & 56.93\% & 91.13\% & \textbf{19.31\%} \\
 & Mistral Large (123B) & \textbf{19.20} & \textbf{13.86} & \textbf{78.59\%} & \textbf{87.83\%} & 19.56\% \\ \hline
\textbf{Closed} & GPT-4o & 14.22 & 11.38 & 54.33\% & 91.67\% & \textbf{20.73\%} \\
 & GPT-o1-mini & \textbf{19.40} & 11.76 & \textbf{87.73\%} & \textbf{86.31\%} & 27.88\% \\
 & Gemini 1.5 Pro & 16.09 & 9.91 & 67.02\% & 89.36\% & 23.25\% \\
 & Claude 3.5 Sonnet & 15.03 & \textbf{13.08} & 58.00\% & 91.25\% & 12.08\% \\ \hline
\textbf{Baseline} & Random Walk & 14.18 & 3.13 & 69.11\% & 89.28\% & 51.25\% \\ \hline
\end{tabular}%
}
\end{table*}
\textbf{Low-parameter LLMs are more invariant to instructions.}
When comparing our task-oriented instruction with the alternative prompts described in Section~\ref{sec:setup}, we find that specifically LLMs with a size of $\leq$ 9 billion parameters maintain significant exploration gaps even when being prompted to focus solely on exploring any novel states. 
For example, Mistral (7B) and Gemma 2 (9B) exhibit exploration gaps of 57\% and 26\% in their last episode. The results suggest that smaller models may face general intrinsic limitations in their ability to explore.
Nevertheless, more competent models on the MMLU-Pro benchmark also do not achieve full coverage. Please find all results comparing the different prompts in Appendix D \cite{grams2025disentangling}.

\begin{figure*}[ht]
    \centering
    \subfigure{
        \centering
        \includegraphics[width=0.48 \linewidth]{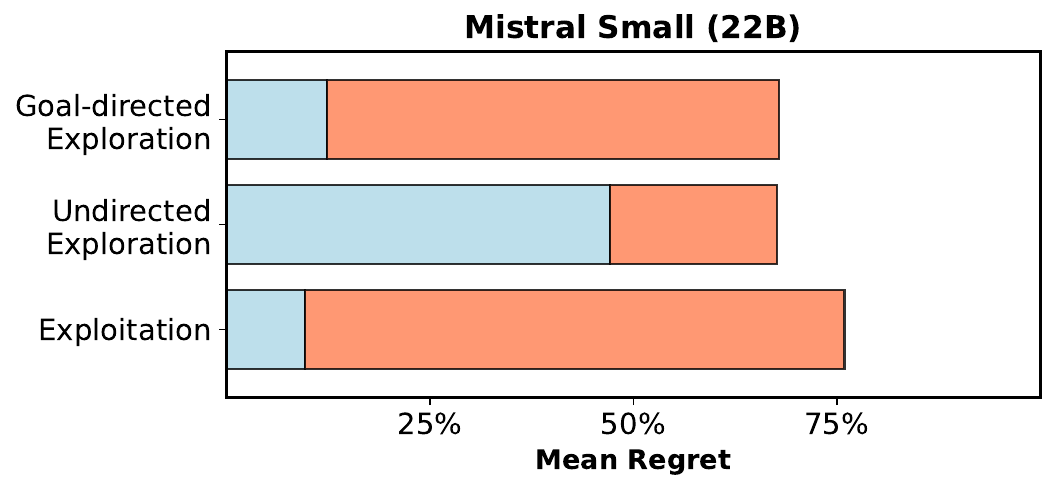}
    }
    \subfigure{
        \centering
        \includegraphics[width=0.48 \linewidth]{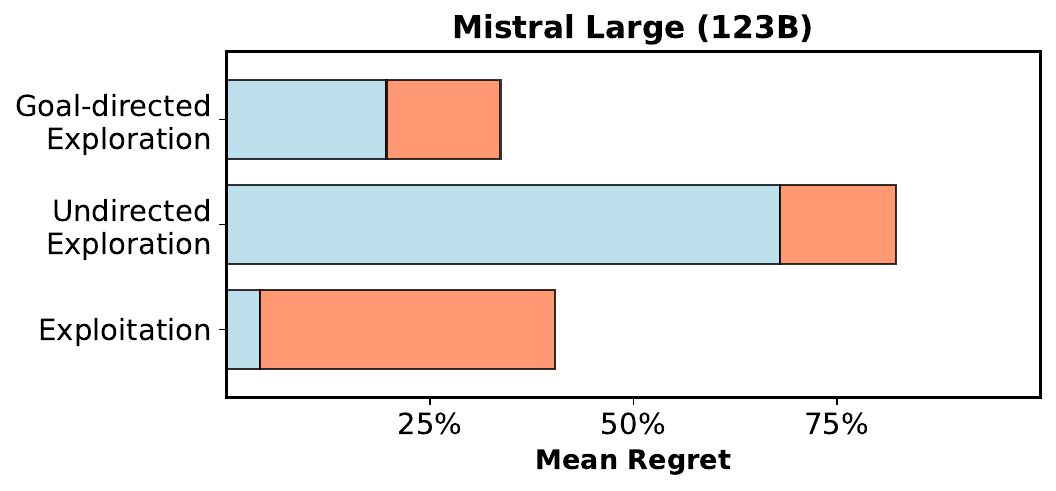}
    }
    \caption{\textbf{The normalized mean regret decomposed into exploration and exploitation components.} Larger LLMs employing task-oriented exploration tend to sacrifice more rewards compared to smaller models. Furthermore, LLMs interpret our exploration instruction differently: For Mistral Small, the gap proportions align more closely with the exploitation prompt, whereas for Mistral Large, they are closer to undirected exploration.}
    \label{fig:main-paper-prompt-breakdown}
\end{figure*}

\textbf{Instructions can have an impact on gaps.}
We observed an increase in the exploitation gap when agents were prompted with an instruction that did not mention the task during exploration. Figure \ref{fig:main-paper-prompt-breakdown} depicts the regret decomposition of runs with the different prompts and Mistral Large. While the LLM maintained a similar task-oriented exploration gap, the behavior significantly differed in terms of agent returns. The undirected exploration instruction leads to more independence across episodes, whereas task-oriented exploration focuses on previously visited regions. On the contrary, when advising the LLM to exploit, then the exploitation gap is constantly minimal.

\textbf{Our decomposition shows differences between models.}
LLMs interpret the same task-oriented exploration instruction with varying behaviors: Lower parameter models predominantly exhibit behavior closer to exploitation when prompted to gather information for the task. In contrast, other models perform more in line with the task-unrelated, intrinsic exploration. Surprisingly, o1-mini achieved lower exploration gaps under task-oriented instructions. It is an interesting question for future research to analyze how compelling intrinsic exploration is for tasks in environments with semantic understanding.

\textbf{Our regret decomposition is a tool for prompt-engineering.}
The results have demonstrated that our decomposition can effectively analyze prompting strategies. We believe it is a valuable tool for refining agent behavior before use in real-world scenarios. Our proposed method represents an initial step towards tools to assess the impact of prompt engineering on the exploration capabilities of LLMs in sequential decision-making. 

\section{Discussion and Future Implications}
Our decomposition and results have several practical and research implications: 1.) Contrary to earlier claims \cite{huang2024wese}, strong models should be used for exploration. We found that models with less reasoning capabilities exhibit significant weaknesses in trajectory discovery. This discrepancy may arise from the increased complexity of finding objects in our setting, while hidden objects in simplified state spaces are easily covered in household domains. 
2.) Agent architectures should be tuned specifically for exploration. Disentangling exploration and exploitation can improve performance in exploration-heavy tasks.
3.) Consistent with prior claims \cite{krishnamurthy2024can, nie2024evolve}, we observe that the most popular LLMs struggle with exploration.
4.) Our experiments have shown that comparing models based on agent returns or another LLM for exploitation does not lead to reliable comparisons.  
Our evaluation procedure fills this gap and allows a fair evaluation of the following promising avenues for future research:

\textbf{Reasoning, self-improvement, and memory.} 
Reasoning, memory, and self-improvement strategies have been proposed. Examples include Reflection \cite{shinn2024reflexion} and RL \cite{guo2025deepseek}. Evaluating how these approaches influence exploration may be a promising research direction. It would also be interesting to see if LLMs are able to self-improve their exploration skills. 

\textbf{Training for intelligent exploration.} 
Training an agent to gather information systematically may be required to develop generalist agents that can act in unknown environments. LLMs should be investigated from a meta-RL perspective. \cite{nie2024evolve} show first evidence that finetuning an agent on exploration trajectories can help in multi-armed bandits. Future agents should be less dependent on prompting and possess intrinsic curiosity to explore their environment. 

\textbf{Small exploration models.}
\hspace{1.5mm}
Improving the ability of weak LLMs to explore is a necessary step towards many real-life applications, requiring embedded and real-time agents. It would also be interesting to develop LLMs that are generally less capable in other domains but can perform strong exploration. Distillation methods could be developed to teach exploration to smaller models. 

\textbf{Improving multi-turn understanding.}
Similarly to \citet{paglieri2024balrog}, we found that agents increasingly produce invalid actions and misunderstand longer sequences of past actions. Developing agents specifically for long-sequence understanding is crucial for future agents. It would be interesting to see if LLMs finetuned on sequential data \cite{zeng2023agenttuning, song2024agentbank} also exhibit better exploration capabilities. 

\section{Conclusion}
We proposed a novel evaluation framework for LLM-based agents that isolates state space exploration through a regret decomposition based on the optimal exploitation return. We argue that developing systems capable of both intelligent exploitation and exploration is a pivotal step toward creating truly generalist agents. We highlighted various research directions for which our method offers a testbed. 

Future work could extend our framework with more complex environments, featuring, for example, randomization and stochasticity. By developing environments that combine complex exploration with underlying rules, domain knowledge, and different tasks, our decomposition can be used to test more intelligent exploration. Benchmarks requiring strategic planning, such as Sokoban. Towers of Hanoi, or BabyAI, could be a starting point. Exploitation and exploration are both complex problems. Therefore, it is essential to continue evaluating them in isolation. 

\begin{ack}
This research was supported in part by the German Federal Ministry for Economic Affairs and Climate Action of Germany (BMWK), and in part by the German Federal Ministry for Research, Technology, and Space (BMFTR).
\end{ack}



\bibliography{library}

\begin{thebibliography}{63}
\providecommand{\natexlab}[1]{#1}
\providecommand{\url}[1]{\texttt{#1}}
\expandafter\ifx\csname urlstyle\endcsname\relax
  \providecommand{\doi}[1]{doi: #1}\else
  \providecommand{\doi}{doi: \begingroup \urlstyle{rm}\Url}\fi

\bibitem[Achiam and Sastry(2017)]{achiam2017surprise}
J.~Achiam and S.~Sastry.
\newblock Surprise-based intrinsic motivation for deep reinforcement learning.
\newblock \emph{arXiv preprint arXiv:1703.01732}, 2017.

\bibitem[Achiam et~al.(2023)Achiam, Adler, Agarwal, Ahmad, Akkaya, Aleman, Almeida, Altenschmidt, Altman, Anadkat, et~al.]{achiam2023gpt}
J.~Achiam, S.~Adler, S.~Agarwal, L.~Ahmad, I.~Akkaya, F.~L. Aleman, D.~Almeida, J.~Altenschmidt, S.~Altman, S.~Anadkat, et~al.
\newblock Gpt-4 technical report.
\newblock \emph{arXiv preprint arXiv:2303.08774}, 2023.

\bibitem[AI(2024{\natexlab{a}})]{mistralAbundance}
M.~AI.
\newblock Mistral small.
\newblock \url{https://mistral.ai/news/september-24-release/}, 2024{\natexlab{a}}.
\newblock [Accessed 10-01-2025].

\bibitem[AI(2024{\natexlab{b}})]{mistralLargeEnough}
M.~AI.
\newblock Mistral large 2i.
\newblock \url{https://mistral.ai/news/mistral-large-2407/}, 2024{\natexlab{b}}.
\newblock [Accessed 10-01-2025].

\bibitem[Anthrophic(2024)]{anthropicIntroducingClaude}
Anthrophic.
\newblock {C}laude 3.5 {S}onnet.
\newblock \url{https://www.anthropic.com/news/claude-3-5-sonnet}, 2024.
\newblock [Accessed 10-01-2025].

\bibitem[Avner et~al.(2012)Avner, Mannor, and Shamir]{avner2012decouplingbandit}
O.~Avner, S.~Mannor, and O.~Shamir.
\newblock Decoupling exploration and exploitation in multi-armed bandits.
\newblock \emph{arXiv preprint arXiv:1205.2874}, 2012.

\bibitem[Bellemare et~al.(2016)Bellemare, Srinivasan, Ostrovski, Schaul, Saxton, and Munos]{bellemare2016unifying}
M.~Bellemare, S.~Srinivasan, G.~Ostrovski, T.~Schaul, D.~Saxton, and R.~Munos.
\newblock Unifying count-based exploration and intrinsic motivation.
\newblock \emph{Advances in neural information processing systems}, 29, 2016.

\bibitem[Burda et~al.(2018)Burda, Edwards, Storkey, and Klimov]{burda2018exploration}
Y.~Burda, H.~Edwards, A.~Storkey, and O.~Klimov.
\newblock Exploration by random network distillation.
\newblock \emph{arXiv preprint arXiv:1810.12894}, 2018.

\bibitem[Chen et~al.(2024)Chen, Zhang, and Zhu]{chen2024efficient}
D.~Chen, Q.~Zhang, and Y.~Zhu.
\newblock Efficient sequential decision making with large language models.
\newblock \emph{arXiv preprint arXiv:2406.12125}, 2024.

\bibitem[Chevalier-Boisvert et~al.(2018)Chevalier-Boisvert, Bahdanau, Lahlou, Willems, Saharia, Nguyen, and Bengio]{chevalier2018babyai}
M.~Chevalier-Boisvert, D.~Bahdanau, S.~Lahlou, L.~Willems, C.~Saharia, T.~H. Nguyen, and Y.~Bengio.
\newblock Babyai: A platform to study the sample efficiency of grounded language learning.
\newblock \emph{arXiv preprint arXiv:1810.08272}, 2018.

\bibitem[Chevalier-Boisvert et~al.(2024)Chevalier-Boisvert, Dai, Towers, Perez-Vicente, Willems, Lahlou, Pal, Castro, and Terry]{chevalier2024minigrid}
M.~Chevalier-Boisvert, B.~Dai, M.~Towers, R.~Perez-Vicente, L.~Willems, S.~Lahlou, S.~Pal, P.~S. Castro, and J.~Terry.
\newblock Minigrid \& miniworld: Modular \& customizable reinforcement learning environments for goal-oriented tasks.
\newblock \emph{Advances in Neural Information Processing Systems}, 36, 2024.

\bibitem[C{\^o}t{\'e} et~al.(2019)C{\^o}t{\'e}, K{\'a}d{\'a}r, Yuan, Kybartas, Barnes, Fine, Moore, Hausknecht, El~Asri, Adada, et~al.]{cote2019textworld}
M.-A. C{\^o}t{\'e}, A.~K{\'a}d{\'a}r, X.~Yuan, B.~Kybartas, T.~Barnes, E.~Fine, J.~Moore, M.~Hausknecht, L.~El~Asri, M.~Adada, et~al.
\newblock Textworld: A learning environment for text-based games.
\newblock In \emph{Computer Games: 7th Workshop, CGW 2018, Held in Conjunction with the 27th International Conference on Artificial Intelligence, IJCAI 2018, Stockholm, Sweden, July 13, 2018, Revised Selected Papers 7}, pages 41--75. Springer, 2019.

\bibitem[Dubey et~al.(2024)Dubey, Jauhri, Pandey, Kadian, Al-Dahle, Letman, Mathur, Schelten, Yang, Fan, et~al.]{dubey2024llama}
A.~Dubey, A.~Jauhri, A.~Pandey, A.~Kadian, A.~Al-Dahle, A.~Letman, A.~Mathur, A.~Schelten, A.~Yang, A.~Fan, et~al.
\newblock The llama 3 herd of models.
\newblock \emph{arXiv preprint arXiv:2407.21783}, 2024.

\bibitem[Ecoffet et~al.(2019)Ecoffet, Huizinga, Lehman, Stanley, and Clune]{ecoffet2019go}
A.~Ecoffet, J.~Huizinga, J.~Lehman, K.~O. Stanley, and J.~Clune.
\newblock Go-explore: a new approach for hard-exploration problems.
\newblock \emph{arXiv preprint arXiv:1901.10995}, 2019.

\bibitem[Grams et~al.(2025)Grams, Betz, Marton, Lüdtke, and Bartelt]{grams2025disentangling}
T.~Grams, P.~Betz, S.~Marton, S.~Lüdtke, and C.~Bartelt.
\newblock Disentangling exploration of large language models by optimal exploitation.
\newblock \emph{arXiv preprint arXiv:2501.08925}, 2025.
\newblock Full version of this paper.

\bibitem[Grigsby et~al.(2023)Grigsby, Fan, and Zhu]{grigsby2023amago}
J.~Grigsby, L.~Fan, and Y.~Zhu.
\newblock Amago: Scalable in-context reinforcement learning for adaptive agents.
\newblock \emph{arXiv preprint arXiv:2310.09971}, 2023.

\bibitem[Guo et~al.(2025)Guo, Yang, Zhang, Song, Zhang, Xu, Zhu, Ma, Wang, Bi, et~al.]{guo2025deepseek}
D.~Guo, D.~Yang, H.~Zhang, J.~Song, R.~Zhang, R.~Xu, Q.~Zhu, S.~Ma, P.~Wang, X.~Bi, et~al.
\newblock Deepseek-r1: Incentivizing reasoning capability in llms via reinforcement learning.
\newblock \emph{arXiv preprint arXiv:2501.12948}, 2025.

\bibitem[Hafner(2021)]{hafner2021benchmarking}
D.~Hafner.
\newblock Benchmarking the spectrum of agent capabilities.
\newblock \emph{arXiv preprint arXiv:2109.06780}, 2021.

\bibitem[Hao et~al.(2023)Hao, Gu, Ma, Hong, Wang, Wang, and Hu]{hao2023worldmodel}
S.~Hao, Y.~Gu, H.~Ma, J.~J. Hong, Z.~Wang, D.~Z. Wang, and Z.~Hu.
\newblock Reasoning with language model is planning with world model.
\newblock \emph{arXiv preprint arXiv:2305.14992}, 2023.

\bibitem[Huang et~al.(2023)Huang, Chen, Mishra, Zheng, Yu, Song, and Zhou]{huang2023selfcorrect}
J.~Huang, X.~Chen, S.~Mishra, H.~S. Zheng, A.~W. Yu, X.~Song, and D.~Zhou.
\newblock Large language models cannot self-correct reasoning yet.
\newblock \emph{arXiv preprint arXiv:2310.01798}, 2023.

\bibitem[Huang et~al.(2024)Huang, Liu, Chen, Wang, Lian, Wang, Tang, and Chen]{huang2024wese}
X.~Huang, W.~Liu, X.~Chen, X.~Wang, D.~Lian, Y.~Wang, R.~Tang, and E.~Chen.
\newblock Wese: Weak exploration to strong exploitation for llm agents.
\newblock \emph{arXiv preprint arXiv:2404.07456}, 2024.

\bibitem[Imani et~al.(2023)Imani, Du, and Shrivastava]{imani2023mathprompter}
S.~Imani, L.~Du, and H.~Shrivastava.
\newblock Mathprompter: Mathematical reasoning using large language models.
\newblock \emph{arXiv preprint arXiv:2303.05398}, 2023.

\bibitem[Jiang et~al.(2023)Jiang, Sablayrolles, Mensch, Bamford, Chaplot, Casas, Bressand, Lengyel, Lample, Saulnier, et~al.]{jiang2023mistral}
A.~Q. Jiang, A.~Sablayrolles, A.~Mensch, C.~Bamford, D.~S. Chaplot, D.~d.~l. Casas, F.~Bressand, G.~Lengyel, G.~Lample, L.~Saulnier, et~al.
\newblock Mistral 7b.
\newblock \emph{arXiv preprint arXiv:2310.06825}, 2023.

\bibitem[Kambhampati et~al.(2024)Kambhampati, Valmeekam, Guan, Verma, Stechly, Bhambri, Saldyt, and Murthy]{kambhampati2024planning}
S.~Kambhampati, K.~Valmeekam, L.~Guan, M.~Verma, K.~Stechly, S.~Bhambri, L.~Saldyt, and A.~Murthy.
\newblock Llms can't plan, but can help planning in llm-modulo frameworks.
\newblock \emph{arXiv preprint arXiv:2402.01817}, 2024.

\bibitem[Ke et~al.(2024)Ke, Sawyer, Soyer, Engelcke, Reichert, Hudson, Reid, Lerchner, Rezende, Lillicrap, et~al.]{ke2024interactive}
N.~R. Ke, D.~P. Sawyer, H.~Soyer, M.~Engelcke, D.~P. Reichert, D.~A. Hudson, J.~Reid, A.~Lerchner, D.~J. Rezende, T.~P. Lillicrap, et~al.
\newblock Can foundation models actively gather information in interactive environments to test hypotheses?
\newblock \emph{arXiv preprint arXiv:2412.06438}, 2024.

\bibitem[Klyubin et~al.(2005)Klyubin, Polani, and Nehaniv]{klyubin2005empowerment}
A.~S. Klyubin, D.~Polani, and C.~L. Nehaniv.
\newblock Empowerment: A universal agent-centric measure of control.
\newblock In \emph{2005 ieee congress on evolutionary computation}, volume~1, pages 128--135. IEEE, 2005.

\bibitem[Krishnamurthy et~al.(2024)Krishnamurthy, Harris, Foster, Zhang, and Slivkins]{krishnamurthy2024can}
A.~Krishnamurthy, K.~Harris, D.~J. Foster, C.~Zhang, and A.~Slivkins.
\newblock Can large language models explore in-context?
\newblock \emph{arXiv preprint arXiv:2403.15371}, 2024.

\bibitem[K{\"u}ttler et~al.(2020)K{\"u}ttler, Nardelli, Miller, Raileanu, Selvatici, Grefenstette, and Rockt{\"a}schel]{kuttler2020nethack}
H.~K{\"u}ttler, N.~Nardelli, A.~Miller, R.~Raileanu, M.~Selvatici, E.~Grefenstette, and T.~Rockt{\"a}schel.
\newblock The nethack learning environment.
\newblock \emph{Advances in Neural Information Processing Systems}, 33:\penalty0 7671--7684, 2020.

\bibitem[Laskin et~al.(2022)Laskin, Wang, Oh, Parisotto, Spencer, Steigerwald, Strouse, Hansen, Filos, Brooks, et~al.]{laskin2022context}
M.~Laskin, L.~Wang, J.~Oh, E.~Parisotto, S.~Spencer, R.~Steigerwald, D.~Strouse, S.~Hansen, A.~Filos, E.~Brooks, et~al.
\newblock In-context reinforcement learning with algorithm distillation.
\newblock \emph{arXiv preprint arXiv:2210.14215}, 2022.

\bibitem[Li et~al.(2021)Li, Kuncoro, Hoffmann, d'Autume, Blunsom, and Nematzadeh]{li2021systematic}
X.~L. Li, A.~Kuncoro, J.~Hoffmann, C.~d.~M. d'Autume, P.~Blunsom, and A.~Nematzadeh.
\newblock A systematic investigation of commonsense knowledge in large language models.
\newblock \emph{arXiv preprint arXiv:2111.00607}, 2021.

\bibitem[Lillicrap(2015)]{lillicrap2015continuous}
T.~Lillicrap.
\newblock Continuous control with deep reinforcement learning.
\newblock \emph{arXiv preprint arXiv:1509.02971}, 2015.

\bibitem[Liu et~al.(2023)Liu, Yu, Zhang, Xu, Lei, Lai, Gu, Ding, Men, Yang, et~al.]{liu2023agentbench}
X.~Liu, H.~Yu, H.~Zhang, Y.~Xu, X.~Lei, H.~Lai, Y.~Gu, H.~Ding, K.~Men, K.~Yang, et~al.
\newblock Agentbench: Evaluating llms as agents.
\newblock \emph{arXiv preprint arXiv:2308.03688}, 2023.

\bibitem[Lu et~al.(2024)Lu, Hu, and Clune]{lu2024intelligent}
C.~Lu, S.~Hu, and J.~Clune.
\newblock Intelligent go-explore: Standing on the shoulders of giant foundation models.
\newblock \emph{arXiv preprint arXiv:2405.15143}, 2024.

\bibitem[Ma et~al.(2023)Ma, Zhang, Wang, Pan, Yu, and Yu]{ma2023laser}
K.~Ma, H.~Zhang, H.~Wang, X.~Pan, W.~Yu, and D.~Yu.
\newblock Laser: Llm agent with state-space exploration for web navigation.
\newblock \emph{arXiv preprint arXiv:2309.08172}, 2023.

\bibitem[Mazoure et~al.(2023)Mazoure, Bruce, Precup, Fergus, and Anand]{mazoure2023accelerating}
B.~Mazoure, J.~Bruce, D.~Precup, R.~Fergus, and A.~Anand.
\newblock Accelerating exploration and representation learning with offline pre-training.
\newblock \emph{arXiv preprint arXiv:2304.00046}, 2023.

\bibitem[Mnih(2013)]{mnih2013playing}
V.~Mnih.
\newblock Playing atari with deep reinforcement learning.
\newblock \emph{arXiv preprint arXiv:1312.5602}, 2013.

\bibitem[Nie et~al.(2024)Nie, Su, Chang, Lee, Chi, Le, and Chen]{nie2024evolve}
A.~Nie, Y.~Su, B.~Chang, J.~N. Lee, E.~H. Chi, Q.~V. Le, and M.~Chen.
\newblock Evolve: Evaluating and optimizing llms for exploration.
\newblock \emph{arXiv preprint arXiv:2410.06238}, 2024.

\bibitem[Norman and Clune(2023)]{norman2023first}
B.~Norman and J.~Clune.
\newblock First-explore, then exploit: Meta-learning intelligent exploration.
\newblock \emph{arXiv preprint arXiv:2307.02276}, 2023.

\bibitem[OpenAI(2024)]{OpenAIo1}
OpenAI.
\newblock {O}pen{A}{I} o1-mini.
\newblock \url{https://platform.openai.com/docs/models#o1}, 2024.
\newblock [Accessed 10-01-2025].

\bibitem[Paglieri et~al.(2024)Paglieri, Cupia{\l}, Coward, Piterbarg, Wolczyk, Khan, Pignatelli, Kuci{\'n}ski, Pinto, Fergus, et~al.]{paglieri2024balrog}
D.~Paglieri, B.~Cupia{\l}, S.~Coward, U.~Piterbarg, M.~Wolczyk, A.~Khan, E.~Pignatelli, {\L}.~Kuci{\'n}ski, L.~Pinto, R.~Fergus, et~al.
\newblock Balrog: Benchmarking agentic llm and vlm reasoning on games.
\newblock \emph{arXiv preprint arXiv:2411.13543}, 2024.

\bibitem[Park et~al.(2024)Park, Liu, Ozdaglar, and Zhang]{park2024regret}
C.~Park, X.~Liu, A.~Ozdaglar, and K.~Zhang.
\newblock Do llm agents have regret? a case study in online learning and games.
\newblock \emph{arXiv preprint arXiv:2403.16843}, 2024.

\bibitem[Pathak et~al.(2017)Pathak, Agrawal, Efros, and Darrell]{pathak2017curiosity}
D.~Pathak, P.~Agrawal, A.~A. Efros, and T.~Darrell.
\newblock Curiosity-driven exploration by self-supervised prediction.
\newblock In \emph{International conference on machine learning}, pages 2778--2787. PMLR, 2017.

\bibitem[Paul(2023)]{paul2023sequential}
S.~K. Paul.
\newblock Sequential planning in large partially observable environments guided by llms.
\newblock \emph{arXiv preprint arXiv:2312.07368}, 2023.

\bibitem[Piriyakulkij et~al.(2024)Piriyakulkij, Langenfeld, Le, and Ellis]{piriyakulkij2024doing}
W.~T. Piriyakulkij, C.~Langenfeld, T.~A. Le, and K.~Ellis.
\newblock Doing experiments and revising rules with natural language and probabilistic reasoning.
\newblock \emph{arXiv preprint arXiv:2402.06025}, 2024.

\bibitem[Ruoss et~al.(2024)Ruoss, Pardo, Chan, Li, Mnih, and Genewein]{ruoss2024lmact}
A.~Ruoss, F.~Pardo, H.~Chan, B.~Li, V.~Mnih, and T.~Genewein.
\newblock Lmact: A benchmark for in-context imitation learning with long multimodal demonstrations.
\newblock \emph{arXiv preprint arXiv:2412.01441}, 2024.

\bibitem[Samvelyan et~al.(2021)Samvelyan, Kirk, Kurin, Parker-Holder, Jiang, Hambro, Petroni, K{\"u}ttler, Grefenstette, and Rockt{\"a}schel]{samvelyan2021minihack}
M.~Samvelyan, R.~Kirk, V.~Kurin, J.~Parker-Holder, M.~Jiang, E.~Hambro, F.~Petroni, H.~K{\"u}ttler, E.~Grefenstette, and T.~Rockt{\"a}schel.
\newblock Minihack the planet: A sandbox for open-ended reinforcement learning research.
\newblock \emph{arXiv preprint arXiv:2109.13202}, 2021.

\bibitem[Sch{\"a}fer et~al.(2021)Sch{\"a}fer, Christianos, Hanna, and Albrecht]{schafer2021decoupling}
L.~Sch{\"a}fer, F.~Christianos, J.~Hanna, and S.~V. Albrecht.
\newblock Decoupling exploration and exploitation in reinforcement learning.
\newblock In \emph{ICML 2021 Workshop on Unsupervised Reinforcement Learning}, 2021.

\bibitem[Schmidhuber(2010)]{schmidhuber2010formal}
J.~Schmidhuber.
\newblock Formal theory of creativity, fun, and intrinsic motivation (1990--2010).
\newblock \emph{IEEE transactions on autonomous mental development}, 2\penalty0 (3):\penalty0 230--247, 2010.

\bibitem[Schmied et~al.(2024)Schmied, Paischer, Patil, Hofmarcher, Pascanu, and Hochreiter]{schmied2024retrieval}
T.~Schmied, F.~Paischer, V.~Patil, M.~Hofmarcher, R.~Pascanu, and S.~Hochreiter.
\newblock Retrieval-augmented decision transformer: External memory for in-context rl.
\newblock \emph{arXiv preprint arXiv:2410.07071}, 2024.

\bibitem[Shinn et~al.(2024)Shinn, Cassano, Gopinath, Narasimhan, and Yao]{shinn2024reflexion}
N.~Shinn, F.~Cassano, A.~Gopinath, K.~Narasimhan, and S.~Yao.
\newblock Reflexion: Language agents with verbal reinforcement learning.
\newblock \emph{Advances in Neural Information Processing Systems}, 36, 2024.

\bibitem[Song et~al.(2024)Song, Xiong, Zhao, Zhu, Wu, Wang, Li, Peng, and Li]{song2024agentbank}
Y.~Song, W.~Xiong, X.~Zhao, D.~Zhu, W.~Wu, K.~Wang, C.~Li, W.~Peng, and S.~Li.
\newblock Agentbank: Towards generalized llm agents via fine-tuning on 50000+ interaction trajectories.
\newblock \emph{arXiv preprint arXiv:2410.07706}, 2024.

\bibitem[Team et~al.(2024{\natexlab{a}})Team, Georgiev, Lei, Burnell, Bai, Gulati, Tanzer, Vincent, Pan, Wang, et~al.]{team2024gemini}
G.~Team, P.~Georgiev, V.~I. Lei, R.~Burnell, L.~Bai, A.~Gulati, G.~Tanzer, D.~Vincent, Z.~Pan, S.~Wang, et~al.
\newblock Gemini 1.5: Unlocking multimodal understanding across millions of tokens of context.
\newblock \emph{arXiv preprint arXiv:2403.05530}, 2024{\natexlab{a}}.

\bibitem[Team et~al.(2024{\natexlab{b}})Team, Riviere, Pathak, Sessa, Hardin, Bhupatiraju, Hussenot, Mesnard, Shahriari, Ram{\'e}, et~al.]{team2024gemma}
G.~Team, M.~Riviere, S.~Pathak, P.~G. Sessa, C.~Hardin, S.~Bhupatiraju, L.~Hussenot, T.~Mesnard, B.~Shahriari, A.~Ram{\'e}, et~al.
\newblock Gemma 2: Improving open language models at a practical size, 2024.
\newblock \emph{URL https://arxiv. org/abs/2408.00118}, 1\penalty0 (3), 2024{\natexlab{b}}.

\bibitem[Valmeekam et~al.(2023)Valmeekam, Marquez, Sreedharan, and Kambhampati]{valmeekam2023planning}
K.~Valmeekam, M.~Marquez, S.~Sreedharan, and S.~Kambhampati.
\newblock On the planning abilities of large language models-a critical investigation.
\newblock \emph{Advances in Neural Information Processing Systems}, 36:\penalty0 75993--76005, 2023.

\bibitem[Wang et~al.(2024)Wang, Ma, Zhang, Ni, Chandra, Guo, Ren, Arulraj, He, Jiang, et~al.]{wang2024mmlu}
Y.~Wang, X.~Ma, G.~Zhang, Y.~Ni, A.~Chandra, S.~Guo, W.~Ren, A.~Arulraj, X.~He, Z.~Jiang, et~al.
\newblock Mmlu-pro: A more robust and challenging multi-task language understanding benchmark.
\newblock In \emph{The Thirty-eight Conference on Neural Information Processing Systems Datasets and Benchmarks Track}, 2024.

\bibitem[Whitney et~al.(2021)Whitney, Bloesch, Springenberg, Abdolmaleki, Cho, and Riedmiller]{whitney2021decoupled}
W.~F. Whitney, M.~Bloesch, J.~T. Springenberg, A.~Abdolmaleki, K.~Cho, and M.~Riedmiller.
\newblock Decoupled exploration and exploitation policies for sample-efficient reinforcement learning.
\newblock \emph{arXiv preprint arXiv:2101.09458}, 2021.

\bibitem[Wu et~al.(2023)Wu, Tang, Mitchell, and Li]{wu2023smartplay}
Y.~Wu, X.~Tang, T.~M. Mitchell, and Y.~Li.
\newblock Smartplay: A benchmark for llms as intelligent agents.
\newblock \emph{arXiv preprint arXiv:2310.01557}, 2023.

\bibitem[Xu et~al.(2023)Xu, Lin, Han, Zhao, Liu, and Cambria]{xu2023logical}
F.~Xu, Q.~Lin, J.~Han, T.~Zhao, J.~Liu, and E.~Cambria.
\newblock Are large language models really good logical reasoners? a comprehensive evaluation from deductive, inductive and abductive views.
\newblock \emph{arXiv preprint arXiv:2306.09841}, 2023.

\bibitem[Yao et~al.(2022{\natexlab{a}})Yao, Chen, Yang, and Narasimhan]{yao2022webshop}
S.~Yao, H.~Chen, J.~Yang, and K.~Narasimhan.
\newblock Webshop: Towards scalable real-world web interaction with grounded language agents.
\newblock \emph{Advances in Neural Information Processing Systems}, 35:\penalty0 20744--20757, 2022{\natexlab{a}}.

\bibitem[Yao et~al.(2022{\natexlab{b}})Yao, Zhao, Yu, Du, Shafran, Narasimhan, and Cao]{yao2022react}
S.~Yao, J.~Zhao, D.~Yu, N.~Du, I.~Shafran, K.~Narasimhan, and Y.~Cao.
\newblock React: Synergizing reasoning and acting in language models.
\newblock \emph{arXiv preprint arXiv:2210.03629}, 2022{\natexlab{b}}.

\bibitem[Yuan et~al.(2023)Yuan, Yuan, Li, Dong, Lu, Tan, Zhou, and Zhou]{yuan2023scaling}
Z.~Yuan, H.~Yuan, C.~Li, G.~Dong, K.~Lu, C.~Tan, C.~Zhou, and J.~Zhou.
\newblock Scaling relationship on learning mathematical reasoning with large language models.
\newblock \emph{arXiv preprint arXiv:2308.01825}, 2023.

\bibitem[Zeng et~al.(2023)Zeng, Liu, Lu, Wang, Liu, Dong, and Tang]{zeng2023agenttuning}
A.~Zeng, M.~Liu, R.~Lu, B.~Wang, X.~Liu, Y.~Dong, and J.~Tang.
\newblock Agenttuning: Enabling generalized agent abilities for llms.
\newblock \emph{arXiv preprint arXiv:2310.12823}, 2023.

\bibitem[Zhou et~al.(2023)Zhou, Xu, Zhu, Zhou, Lo, Sridhar, Cheng, Ou, Bisk, Fried, et~al.]{zhou2023webarena}
S.~Zhou, F.~F. Xu, H.~Zhu, X.~Zhou, R.~Lo, A.~Sridhar, X.~Cheng, T.~Ou, Y.~Bisk, D.~Fried, et~al.
\newblock Webarena: A realistic web environment for building autonomous agents.
\newblock \emph{arXiv preprint arXiv:2307.13854}, 2023.

\end{thebibliography}

\end{document}